\documentclass[manuscript,authorversion,nonacm]{acmart}

\AtBeginDocument{%
  }

\author{Hellina Hailu Nigatu}

\email{hellina_nigatu@berkeley.edu}
\affiliation{%
  \institution{University of California at Berkeley}
  \city{Berkeley}
  \state{CA}
  \country{USA}
}
\author{ Zeerak Talat}

\email{z@zeerak.org}
\affiliation{%
  \institution{Edinburgh University}
  \state{Edinburgh}
  \country{UK}
}

\usepackage{natbib}
\begin{document}

\title{A Capabilities Approach to Studying Bias and Harm in Language Technologies}



\maketitle
\section{Introduction}
Mainstream Natural Language Processing (NLP) research has ignored the majority of the world's languages.
In moving from excluding the majority of the world's languages to blindly adopting what we make for English, we first risk importing the same harms we have at best mitigated and at least measured for English. 
For instance, \citet{Yong2023Oct} showed how prompting GPT-4 in low-resource languages circumvents guardrails that are effective in English.
However, in evaluating and mitigating harms arising from adopting new technologies into such contexts, we often disregard (1) the actual community needs of Language Technologies, and (2) biases and fairness issues within the context of the communities.

Here, we 
consider fairness, bias, and inclusion in Language Technologies 
through the lens of the Capabilities Approach~\cite{Robeyns2011Apr}. 
The Capabilities Approach centers \textit{what people are capable of achieving}, given their intersectional social, political, and economic contexts instead of \textit{what resources are (theoretically) available to them.}
In the following sections, we 
detail 
the Capabilities Approach, 
its relationship to multilingual and multicultural evaluation, and how the framework affords meaningful collaboration with community members in defining and measuring harms of Language Technologies.






\section{The Capabilities Approach} 
The Capabilities Approach is a framework in developmental economic studies proposed by Amartya Sen in a series of articles published as far back as 1974~\cite{Sen1994}. It has been applied to varied fields including environmental justice \cite[e.g.][]{10.1080/19452829.2014.899563} and technological design \cite[e.g.][]{10.1080/19452829.2011.576661}. 
The framework posits that well-being should be understood in terms of people's capabilities and functionings, given that the freedom to achieve well-being is of primary moral importance. 
Capabilities refer to ``doings and beings that people can achieve if they choose.'' and functionings refer to ``capabilities that have been realized.'' 
The core of the framework is that knowing what resources a person owns or can use is insufficient to assess what they can achieve.
The capabilities approach puts our focus on the \textit{means} that would lead to a valuable end and stresses that there are multiple means to the same end. 
%
%
The framework 
centers human diversity by accepting 
the plurality of functionings and capabilities as important evaluative spaces, explicitly focusing on personal and socio-environmental conversion factors\footnote{Conversion factors are factors that affect to what extent a capability can be realized to a functioning. For example, a disability that might hinder someone from using public transport is a conversion factor in whether or not they can use the resource even if it is available to them.}, and acknowledging human agency and diversity of goals~\cite{Robeyns2011Apr}. 

\section{The Majority World}
The ``Majority World'' refers to the peoples of African, Indigenous, Asian, and Latin American descent. 
The term was coined by \citet{Alam2024} to emphasize that these populations constitute the majority of the world's population but are continually referred to with marginalizing terminology. 
However, the Majority world continues to suffer exploitation and marginalization~\cite{Birhane2020Aug}. 
Within 
NLP research, 
they experience harm and marginalization (1) by being excluded from research and development of many of the State-Of-The-Art models~\cite{Ojo2023Nov}, (2) through direct exploitation of their labor for training and moderating models~\cite{Musambi2023Jun}, and (3) via harm caused by failures in existing models~\cite{nigatu_facct}. 
Under this umbrella term lie several cultures, languages, and communities all with varying levels of economic, political, and social fabrics, which makes evaluation of bias in these contexts difficult~\cite{talat-etal-2022-reap}. 
In moving towards an inclusive evaluation ecosystem, an important question to ask is: \textbf{how do we make the diverse and entangled identities and social fabrics of the `Majority World' explicit in our evaluation?}

While there are several approaches to realizing the capabilities approach into a theory, here we propose using \citeauthor{Sen1994}'s 
procedural approach~\cite{Professor2005Jul}. 
The procedural approach emphasizes community ownership over determining what is important and avoids researchers putting subjective constraints.
Some questions to consider when applying the capabilities approach to a specific community are: 

\begin{quote}
    \textbf{Q1:} Does this community have the capability to use this resource (Language Technology)?

    \begin{itemize}
        \item \textbf{Q1.1}. What limits the community from choosing to realize the capability to a functioning?
        \item \textbf{Q1.2}. What conversion factors do the community members have at their disposal?
        \item \textbf{Q1.3}. What is the list of capabilities (both individual and community level) the community members have in realizing these resources into a functioning?
    \end{itemize}
    
  \textbf{Q2}. What other resources (Language Technologies) does the community have the capability to utilize?
  \begin{itemize}
      \item \textbf{Q2.1}. What are the community needs?
      \item \textbf{Q2.2}. What other forms of harm is the community dealing with that we are not measuring? 
  \end{itemize}
    
\end{quote}


In addition to ignoring the majority of the world's population from research for centuries, the academic community, which is largely WEIRD\footnote{WEIRD is an acronym for Western, Educated, Industrialized, Rich, and Democratic. 
The term has been used by prior work to demonstrate the hegemony in different fields \cite[e.g.][]{10.1145/3593013.3593985, Henrich2010Jun}}, has maintained a 
tradition of extractive parachute studies~\cite{Abebe2021NarrativesAC}. 
Hence, when thinking about ways of building inclusive evaluation structures, meaningful community engagement should be central in our reasoning. 
The capabilities approach provides a shift from ```What resources have we allocated and how can we measure its impact?'' to ``What resources do community members have the capability to utilize?'' thereby centering the needs of the community in question.
Additionally, the very definition of the approach rests in individuals having real freedoms and choices. By applying the capabilities approach, we can conduct evaluations that center the agency and the needs of community members.
\bibliographystyle{ACM-Reference-Format}
\bibliography{custom} 

\end{document}